\documentclass{article}

\usepackage[final]{nips_2017}

\usepackage[utf8]{inputenc} 
\usepackage[T1]{fontenc}    
\usepackage{hyperref}       
\usepackage{url}            
\usepackage{booktabs}       
\usepackage{amsfonts}       
\usepackage{nicefrac}       
\usepackage{microtype}      
\usepackage{amsmath}
\usepackage{todonotes}{}
\usepackage{pgfplots}
\pgfplotsset{compat=1.14}
\title{Arbitrary Discrete Sequence Anomaly Detection with Zero Boundary LSTM}
\author{
Chase Roberts\thanks{Work done while an intern at Bloomberg}\\Rensselaer Polytechnic Institute\\
\texttt{roberc4@rpi.edu}
\And Manish Nair \\Bloomberg \\ \texttt{mnair4@bloomberg.net}
}
\begin{document}
\maketitle

\begin{abstract}
We propose a simple mathematical definition and new neural architecture for finding anomalies within discrete sequence datasets.
Our model comprises of a modified LSTM autoencoder and an array of One-Class SVMs.
The LSTM takes in elements from a sequence and creates context vectors that are used to predict the probability distribution of the following element.
These context vectors are then used to train an array of One-Class SVMs.
These SVMs are used to determine an outlier boundary in context space.
We show that our method is consistently more stable and also outperforms standard LSTM and sliding window anomaly detection systems on two generated datasets.
\end{abstract}

\section{Introduction}
Discrete sequence anomaly detection is the process of discovering sequences in a dataset that do not conform
to the pattern seen in a normal dataset. 
Detecting anomalies in discrete sequence data, especially text or byte datasets,
is an incredibly difficult but valuable problem. Discovering these anomalies can help in many domains such as cybersecurity [3] and flight safety [4]. 
However, implementing these detection systems has proven to be a challenge. 
This is mainly because formally defining what makes a sequence anomalous is a non-trivial task. 
As such, metrics for an algorithm's success becomes ill-defined, mainly relying on human validation.
In this paper, we give a theoretical definition of anomalies for discrete sequences and develop 
a machine learning architecture that is derived directly from this definition.
\subsection{Prior Work}
Existing discrete sequence anomaly detection systems are comprised of naive sliding window, Markov chain, or 
subsequence frequency analysis [1]. While these systems are effective in catching local anomalies within a sequence, they tend to to miss long-term dependencies on more structured sequences. Errors like grammatical mistakes in sentences may require looking back several dozen characters to determine an anomaly. Increasing window sizes for an n-gram model will either require significantly more training data or cause a huge increase in false positives.

Recently, more modern deep learning approaches have consisted of training
an LSTM to predict the probability distribution of the next element in a sequence [14], and flagging that
sequence as an anomaly if the predicted probability is below a certain threshold. 
However, if we use standard minibatch and cross entropy loss to train our LSTM, then the predict variance of infrequent values becomes quite large. This makes deciding the threshold for anomalies very difficult, and grays out the boundary between real and anomalous sequences. The instability of this decision bound makes this system difficult to use as an alerting system, as it either raises too many false positive alerts or misses an unacceptable number of true positives.
\section{Background}

\subsection{Autoencoders}
Autoencoders are an unsupervised way of decreasing the dimensionality of a given input sequence. 
They have been shown to outperform other dimensionality reduction techniques, such as PCA, for feature extraction [15]. Researchers have used autoencoders in the past for anomaly detection [5], image generation [6], and various input interpolation, including sentence interpolation [7]. \\ 

Autoencoders work by training two functions: the ``encoder'' $Enc(x): \mathbb{R}^n \rightarrow \mathbb{R}^m$ and the ``decoder'' $Dec(x): \mathbb{R}^m \rightarrow \mathbb{R}^n$ where $n >> m$. These two functions are trained simultaneously by decreasing the loss function $L = d(Dec(Enc(x)), x)$ where $d$ is some reconstruction distance metric such as mean squared error or cross entropy. By forcing the neural network to reconstruct its own input through a bottleneck, the vector at the bottleneck becomes a low-dimensional rich description of the input data. \\ 

Past methods of anomaly detection using autoencoders would use the reconstruction loss as the metric for detection [5]. However, Vincent [8] showed that the reconstruction loss is actually equal to the gradient of the energy function for the underlying probability distribution. This causes certain anomalous inputs to be misclassified as normal inputs if they happen to have a near zero gradient. Energy-based models [9] have been developed to correct for this, but these models do not extend to discrete sequences.
\subsection{One Class SVMs}
One-Class SVMs (OCSVMs) [10] are density boundary estimators for fixed space inputs. Given a set of training vectors $\mathbf{X}$ and a value $\nu$ between 0 and 1, OCSVMs are able to create a boundary where a lower bound of $1 - \nu$ of the vectors will be given a positive value and an upper bound of $\nu$ of the vectors will be given a negative value. Any input that is assigned a value less than 0 will be considered an outlier or anomalous input. \\

Training a OCSVM requires solving the quadratic programming problem:
\[
\min_{A}\sum_{ij}\alpha_i\alpha_jK(x_i, x_j)\  \text{subject to}\  0 \leq \alpha_i \leq \frac{1}{\nu l},
\sum_i \alpha_i = 1
\]
where $A$ is a vector of scalars $\alpha \in A$, $K$ is the Gaussian kernel $K(x, y) = e^{|x - y|^2/\gamma}$, $x \in \mathbf{X}$ are our datapoints , and $l = |\mathbf{X}|$. The support vector expansion is then:
\[
SV(y) = \sum_i \alpha_iK(x_i, y) - \rho
\]
where $\rho = \sum_j \alpha_j K(x_j, x_i)$ for any given $i$ where $\alpha_i \approx 0$. The decision function is usually the sign of $SV(y)$. However, for our method we will be using a variable cutoff that may not be zero.
\section{Proposed Method}
\subsection{Theoretical Foundation}
Let $\Sigma$ represent our finite alphabet. 
Let $\mathbf{x} \in \Sigma^n$ represent a discrete sequence (we implicitly assume that each of our sequences start and end with a special delimiter element).
Let $\mathbf{L} = \{\mathbf{x}_1, \mathbf{x}_2,...,\mathbf{x}_m\}$ be the language of our system. We define $\mathbf{x}$ as an anomaly simply if:
\[
\mathbf{x} \not \in \mathbf{L}
\]
Equivalently, it also follows that $\mathbf{x}$ is a anomaly if there exists $x_i \in \mathbf{x}$ such that: 
\[P(x_i | x_1, ..., x_{i-1}; \mathbf{L}) = 0\] 
From this definition, we wish to build a neural network architecture to approximate the function $f(x; \mathbf{L})$ where: 
\[
    f(\mathbf{x}; \mathbf{L})= 
\begin{cases}
    1,& \text{if } \exists x_i \in \mathbf{x}\ :\ P(x_i | x_1, ..., x_{i-1}; \mathbf{L}) = 0\\
    0,              & \text{otherwise}
\end{cases}
\]

Since we require a probability of zero, it follows that:
\[
P(x_i | x_1, ..., x_{i-1};\mathbf{L}) = P(x_1, ..., x_{i-1}| x_i;\mathbf{L}) = 0
\]
Because we have now flipped the conditional of the probability distribution, we can now use a separate probability distribution for each element in our alphabet. Because of this, $f(\mathbf{x})$ is equivalently:
\[
    f(\mathbf{x}; \mathbf{L})= 
\begin{cases}
    1,& \text{if } \exists x_i \in \mathbf{x}\ :\ P_{x_i}(x_1, ..., x_{i-1}; \mathbf{L}) = 0\\
    0,              & \text{otherwise}
\end{cases}
\]
where $P_{x}$ is the probability distribution under a specified byte $x$. If we are able to reduce the input for our probability distribution $\{x_1, ..., x_{i-1}\}$ into a vector of fixed size, then we are able to use standard density estimators for $P_x$. In the following section, we describe exactly how we achieved this.
\subsection{Zero Boundary LSTM}
\begin{figure}[h]
  \centering
  \includegraphics[scale=0.4]{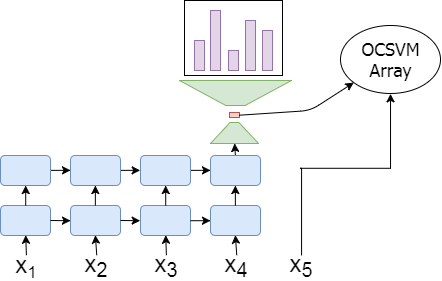}
  \caption{A diagram of our neural network architecture. Here, to detect if element $x_5$ is an 
    anomaly, we calculate the context vector up to $x_5$ and pass it into the corresponding OCSVM.}
\end{figure}
We define:  
\begin{align*}
  E(\mathbf{x};\theta_E) &: \Sigma^n \rightarrow \mathbb{R}^{n \times e}\\
  D(y;\theta_D) &: \mathbb{R}^{e} \rightarrow \mathbb{R}^{|\Sigma|} \\
  O_{\sigma}(cv) &: \mathbb{R}^{e} \rightarrow \mathbb{R} \\
\end{align*}
where $E(\mathbf{x};\theta_E)$ is an LSTM encoder, $D(y;\theta_D)$ is a MLP decoder, and $O_\sigma(cv)$ is the OCSVM for element $\sigma \in \Sigma$, $n$ is the length of the sequence $\mathbf{x}$, and $e$ is the size of our bottleneck. We define the OCSVM array as $\mathbf{O} = \{ O_\sigma \forall \sigma \in \Sigma\}$. We also define $E(\mathbf{x}; \theta_E)_i$ as the $i$th row of $E(\mathbf{x}; \theta_E)$. This vector will also be referenced as a context vector, as it encodes the context up to the $i$th element.  \\ 

Normal LSTM autoencoders are trained by reconstructing their own input. However, as a modification, our method outputs the expected probability distribution for the $i+1$ element given the first $i$ elements.
Thus, we can then learn $\theta_E$ and $\theta_D$ by using stochastic gradient decent where our loss is the cross entropy loss of the predicted character.
\[
\mathcal{L} = \sum_{i=1}^{n-1}-\log P(x_{i+1}|D(E(\mathbf{x}; \theta_E)_i;\theta_D))
\]

We will now describe how to train the OCSVM array. Let $\theta_E'$ and $\theta_D'$ be pretrained parameters. Let $\mathbf{X} = \{\mathbf{x_1},...,\mathbf{x_N}\} \subseteq \mathbf{L}$ be our training set. Let $\mathbf{Y} = \{Y_\sigma\ \forall \sigma \in \Sigma\}$ where $Y_\sigma$ is a set of context vectors such that:
\[
Y_\sigma = \{ E(\mathbf{x}; \theta_E')_i : x_{i} = \sigma\ \forall \mathbf{x} \in \mathbf{X} \}
\]
We can then train each OCSVM $O_\sigma \in \mathbf{O}$ with the context vector set from the corresponding $Y_\sigma \in \mathbf{Y}$ using the method from Schölkopf [10].\\

We now describe how to approximate the function $f(\mathbf{x})$. Given, $\theta_E'$ and $\mathbf{O}$, we define $g \approx f$ as
\[
    g(\mathbf{x})= 
\begin{cases}
    1,& \text{if } \exists x_i \in \mathbf{x} :O_{x_i}(E(\mathbf{x}; \theta_E')_{i-1})  < t_\sigma\\
    0,              & \text{otherwise}
\end{cases}
\]
where $t_\sigma$ is a threshold hyperparameter used to control the Type I error for each OCSVM. The value tends to be slightly less than zero.\\

The motivation behind our architecture comes from Heller et al. [2]. They stated that the failure of their window registry anomaly detection came from the fact that their kernel was unable to incorporate prior distributions of features. By pretraining our LSTM autoencoder to predict the distribution of the following element, the vector at the bottleneck now incorporates all of our prior information of both the sequence and the data distribution.
\section{Experiments}

We compare our method against two other anomaly detection algorithms: standard LSTM next character prediction and naive sliding window. Our experiments consist of two generated toy datasets. The toy datasets are generated IPv4 addresses and generated string only JSON objects.\\ 

Our LSTMs were implemented using tensorflow [11] and we used the scikit learn implementation of OCSVM [12]. Our OCSVMs were all trained with $\nu = 0.001$ and the remaining parameters were left as their defaults. 
For both of our LSTMs, we used an embedding size of 128 for each of the elements in our alphabet. Our LSTMs used 5 layered cells with 128 hidden units in each layer. 
For the Zero Boundary LSTM, we used an MLP bottleneck with seven hidden layers of shape $\{128, 128, 64, 32, 64, 128, 256\}$. Each layer used the ReLU activation function except for the bottleneck layer which used linear activation. The standard LSTM used as a baseline has the same architecture and training procedure as the Zero Boundary LSTM, except we use two hidden layers of size 256 rather than a bottleneck. Both LSTMs used a logits of size 257 (256 possible bytes + 1 for special start/end of sequence element). We trained both networks with the Adam optimizer [16] with default parameters. The cutoff used for anomaly detection is variable and we explicitly state the cutoff used for each experiment. Our naive sliding window used a different window size for each experiment and we only publish on the window size that had the best results.

\subsection{IPv4}
In our first experiment, all of our models were trained on 10,000 randomly generated IPv4 addresses. These strings were generated by selecting four random integers between 0 - 255 uniformly and placing a dot between each number. We then created four different classes of anomalies to try to detect: trivial, length, digit, and dot placement. Trivial anomalies are anomalies that contain characters other than a digit or a dot. Length anomalies are anomalies where the IP address was either cut off too soon or extends for more than four digit groupings. Digit anomalies are strings that contain digit groupings of values larger than 255. Finally, dot placement anomalies are anomalies where a string either contains a double dot (for example: ``123.123..123.123'') or either starts or ends with a dot.\\

The $t_\sigma$ values of our Zero Boundary LSTM were set to the lowest value calculated from our training and validation sets. The cutoff for the standard LSTM was set to be 1 in 8103. Both LSTMs were trained with two epochs of the training data. The n-gram model used a window of size four. The results for this experiment can be seen in Figure 2.
\pgfplotstableread[row sep=\\,col sep=&]{
Count    & Zero & Normal & n-gram-4 \\ 
Trivial  & 1000          & 1000   & 1000    \\
Digit    & 951           & 912    & 1000    \\
Length   & 1000          & 919    & 65      \\
Dot      & 1000          & 922    & 1000    \\
Test Set & 1000          & 986    & 991     \\ 
    }\ipresults
\begin{figure}[h]
  \centering

  \begin{tikzpicture}
      \begin{axis}[
              width=.85\textwidth,
              height=.4\textwidth,
              legend style={at={(1.01,.8)},
                  anchor=west,legend columns=1},
              ybar,
              symbolic x coords={Trivial,Digit,Length,Dot,Test Set},
              xtick=data,
              ylabel=Count,
              xlabel=Dataset Type,
              title=IPv4s,
          ]
          \addplot table[x=Count,y=Zero]{\ipresults};
          \addplot table[x=Count,y=Normal]{\ipresults};
          \addplot table[x=Count,y=n-gram-4]{\ipresults};
          \legend{Zero Boundary, Normal LSTM, n-gram-4}

      \end{axis}
  \end{tikzpicture}
      \caption{Experimental results with our IPv4 dataset. Here, we see that our Zero Boundary LSTM outperforms the standard LSTM algorithm in all categories. The Zero boundary does not perform as well as the n-gram model for anomalous digit dataset, as the ngram has the capacity to memorize all possible acceptable numbers. However, due to the window size being so short, n-grams can not detect anomalies where there are too many digit groupings, unlike the Zero Boundary and normal LSTMs.}
\end{figure}
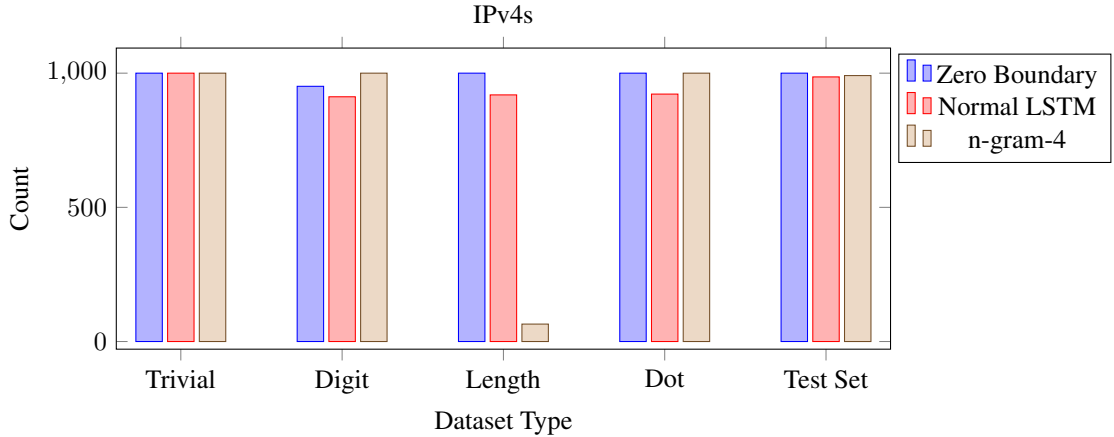
\subsection{JSONs}
The next toy dataset experiment used randomly generated JSON strings. These strings contain nested objects up to four levels deep. Each object contains between zero and four random entries. The entries for any given JSON object are either lowercase strings with lowercase string keys, or nested JSON objects with lowercase string keys. The probability of nesting for any given entry was set to be 1 in 5. We then created four different classes of anomalies to detect: colon, comma, quote, and nesting. Colon anomalies are anomalies where a colon is misplaced. Comma anomalies are when commas are either incorrectly placed or missing. Quote anomalies are where quote markers are either misplaced or completely missing. Nesting anomalies are when the curly braces do not add up correctly. \\

In this experiment, we trained both LSTMs for four epochs of the data. We used an n-gram window size of three. The results of this experiment can be seen in Figure 3. Our Zero Boundary LSTM was able to nearly match the performance of the normal LSTM. However, these results are from highest performing models after many attempts and hyper parameter searches. One of the main advantages of our system is the stability we see with training. \\

In the next experiment, we showcased the models trained with different epoch values. As you can see in Figure 4, the Zero Boundary LSTM has a much easier time consistently finding a stable decision boundary than the normal LSTM. 

\pgfplotstableread[row sep=\\,col sep=&]{
Count    & Zero & Normal & n-gram-3 \\
Colon    & 1000          & 1000    & 977     \\
Comma    & 1000          & 1000    & 948     \\
Nesting  & 994           & 1000    & 991     \\
Quote    & 1000          & 1000    & 1000    \\
Test Set & 998           & 1000    & 1000   \\
    }\jsonresults
\begin{figure}[h]
  \centering

  \begin{tikzpicture}
      \begin{axis}[
              width=.85\textwidth,
              height=.4\textwidth,
              legend style={at={(1.01,.8)},
                  anchor=west,legend columns=1},
              ybar,
              symbolic x coords={Colon,Comma,Nesting,Quote,Test Set},
              xtick=data,
              ylabel=Count,
              xlabel=Dataset Type,
              title=JSONs,
          ]
          \addplot table[x=Count,y=Zero]{\jsonresults};
          \addplot table[x=Count,y=Normal]{\jsonresults};
          \addplot table[x=Count,y=n-gram-3]{\jsonresults};
          \legend{Zero Boundary, Normal LSTM, n-gram-3}

      \end{axis}
  \end{tikzpicture}
      \caption{Experimental results with our JSON dataset. Results are from the best models that we could train.}
\end{figure}
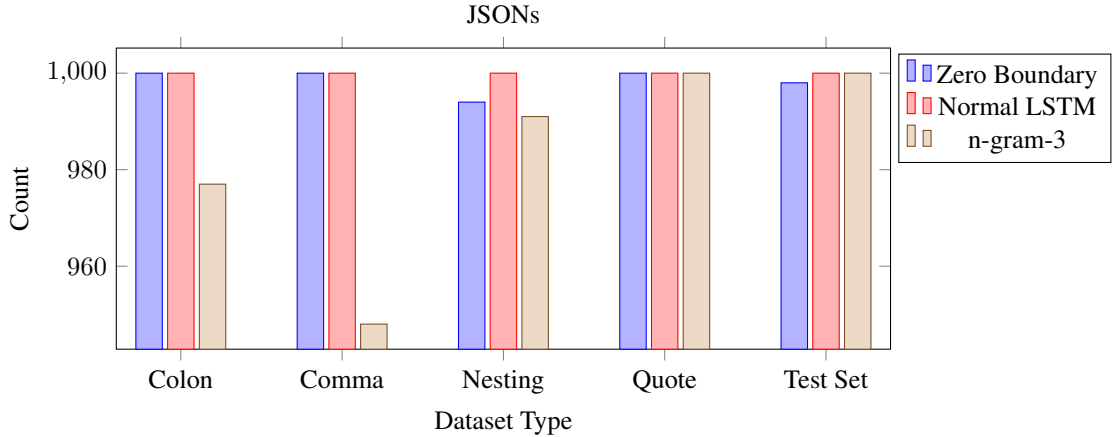

\pgfplotstableread[row sep=\\,col sep=&]{
Counts  & A & B & C & D \\
Epoch 1 & .998                       & .998                        & .363                & .811                 \\
Epoch 2 & .998                       & .995                        & .998                & 1                    \\
Epoch 3 & .998                       & 1                           & .62                 & .564                  \\
Epoch 4 & .998                       & 1                           & 1                   & 1                    \\
    }\constresults
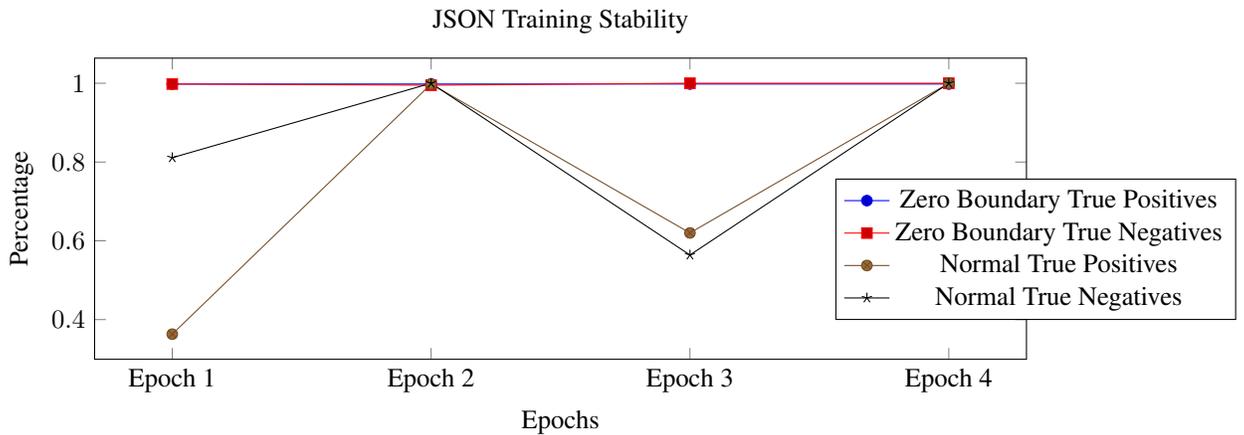
\begin{figure}[h]
  \centering

  \begin{tikzpicture}
      \begin{axis}[
              width=\textwidth,
              height=.4\textwidth,
              legend style={at={(1.01,.6)},
                  anchor=north,legend columns=1},
              symbolic x coords={Epoch 1, Epoch 2, Epoch 3, Epoch 4},
              xtick=data,
              ylabel=Percentage,
              xlabel=Epochs,
              title=JSON Training Stability,
          ]
          \addplot table[x=Counts,y=A]{\constresults};
          \addplot table[x=Counts,y=B]{\constresults};
          \addplot table[x=Counts,y=C]{\constresults};
          \addplot table[x=Counts,y=D]{\constresults};
          \legend{Zero Boundary True Positives, Zero Boundary True Negatives, Normal True Positives, Normal True Negatives}

      \end{axis}
  \end{tikzpicture}
      \caption{Training stability results from our Zero Boundary LSTM and normal LSTM models. Accuracies are determined from the same datasets used for the other JSONs experiment.}
\end{figure}
\section{Future Work}
One of the largest drawbacks to our system is its intolerance to having anomalies within our training set. To get around this, we use a $\nu$ value that is not zero. However, this workaround also tends to increase false positive rating, especially for sequences that have context vectors right on the edge of the decision bound. Also, using OCSVMs as our density estimators makes scaling to larger datasets difficult, as OCSVM have a training run time of $O(n^3)$. Solving these two issues will be a great improvement to our system and may even make unsupervised anomaly detection in areas like network traffic finally viable. \\ 

\section{Conclusion}
In this paper, we developed a mathematical definition of anomalies for discrete sequence datasets and created a neural network architecture to approximate this definition. Our method is able to give a stable decision boundary that does not suffer from the variance of SGD training while also outperforming or matching the performance of both the sliding window and standard unsupervised LSTM algorithms. Even with these successes, our system still suffers heavily when scaled to very large datasets due to the cubic nature of OCSVMs. In order to use this system for more practical situations, this will need to be solved.
\subsubsection*{Acknowledgments}

We would like to thank Raphael Meyer, Connie Yee and Sheryl Zhang of Bloomberg for their valuable 
contributions to this work.

\section*{References}

\small

[1] Chandola, Varun, Arindam Banerjee, and Vipin Kumar. "Anomaly detection: A survey." ACM computing surveys (CSUR) 41.3 (2009): 15.

[2] Heller, Katherine A., et al. "One class support vector machines for detecting anomalous windows registry accesses." Proc. of the workshop on Data Mining for Computer Security. Vol. 9. 2003.

[3] Wang, Ke, and Salvatore J. Stolfo. "Anomalous payload-based network intrusion detection." RAID. Vol. 4. 2004.

[4] Budalakoti, Suratna, Ashok N. Srivastava, and Ram Akella. "Discovering atypical flights in sequences of discrete flight parameters." Aerospace Conference, 2006 IEEE. IEEE, 2006.

[5] An, Jinwon, and Sungzoon Cho. Variational Autoencoder based Anomaly Detection using Reconstruction Probability. Technical Report, 2015.

[6] Tieleman, Tijmen. Optimizing neural networks that generate images. Diss. University of Toronto (Canada), 2014.

[7] Bowman, Samuel R., et al. "Generating sentences from a continuous space." arXiv preprint arXiv:1511.06349 (2015).

[8] Vincent, Pascal. "A connection between score matching and denoising autoencoders." Neural computation 23.7 (2011): 1661-1674.

[9] Zhai, Shuangfei, et al. "Deep structured energy based models for anomaly detection." International Conference on Machine Learning. 2016.

[10] Schölkopf, Bernhard, et al. "Support vector method for novelty detection." Advances in neural information processing systems. 2000.

[11] Abadi, Martín, et al. "Tensorflow: Large-scale machine learning on heterogeneous distributed systems." arXiv preprint arXiv:1603.04467 (2016).

[12] Pedregosa, Fabian, et al. "Scikit-learn: Machine learning in Python." Journal of Machine Learning Research 12.Oct (2011): 2825-2830.

[13] Yu, Lantao, et al. "SeqGAN: Sequence Generative Adversarial Nets with Policy Gradient." AAAI. 2017.

[14] Nanduri, Anvardh, and Lance Sherry. "Anomaly detection in aircraft data using Recurrent Neural Networks (RNN)." Integrated Communications Navigation and Surveillance (ICNS), 2016. IEEE, 2016.

[15] Huang, Fu Jie, Y-Lan Boureau, and Yann LeCun. "Unsupervised learning of invariant feature hierarchies with applications to object recognition." Computer Vision and Pattern Recognition, 2007. CVPR'07. IEEE Conference on. IEEE, 2007.

[16] Kingma, Diederik, and Jimmy Ba. "Adam: A method for stochastic optimization." arXiv preprint arXiv:1412.6980 (2014).
\end{document}